\documentclass{article}

\usepackage[english]{babel}
\usepackage{subcaption}

\usepackage[letterpaper,top=2cm,bottom=2cm,left=3cm,right=3cm,marginparwidth=1.75cm]{geometry}
\usepackage{mathtools, nccmath}
\usepackage{amssymb, amsthm, mathrsfs}
\usepackage{stmaryrd}
\usepackage{lipsum} 
\usepackage{multirow}
\usepackage{booktabs}
\usepackage{amsmath}
\usepackage{graphicx}
\usepackage[colorlinks=true, allcolors=blue]{hyperref}
\usepackage{tabularx}
\usepackage{rotating}
\usepackage{float}

\title{Improving the Performance of Radiology Report De-identification with Large-Scale Training and Benchmarking Against Cloud Vendor Methods}
\author{
Eva Prakash\\
Stanford University
\and
Maayane Attias\\
JP Morgan Chase \& Co
\and
Pierre Chambon\\
Sorbonne University
\and
Justin Xu\\
University of Oxford
\and
Steven Truong\\
NVIDIA
\and
Jean-Benoit Delbrouck\\
HOPPR
\and
Tessa Cook\\
University of Pennsylvania
\and
Curtis Langlotz\\
Stanford University\\
}
\date{}
\begin{document}

\maketitle

\vspace{-2em} 

\section{Abstract}\label{abstract}
\textbf{Objective}: To enhance automated de-identification of radiology reports by scaling transformer-based models through extensive training datasets and benchmarking performance against commercial cloud vendor systems for protected health information (PHI) detection.
\\\textbf{Materials and Methods}: In this retrospective study, we built upon a state-of-the-art, transformer-based, PHI de-identification pipeline by fine-tuning on two large annotated radiology corpora from Stanford University, encompassing chest X-ray, chest CT, abdomen/pelvis CT, and brain MR reports and introducing an additional PHI category (AGE) into the architecture. Model performance was evaluated on test sets from Stanford and the University of Pennsylvania (Penn) for token-level PHI detection. We further assessed (1) the stability of synthetic PHI generation using a “hide-in-plain-sight” method and (2) performance against commercial systems. Precision, recall, and F1 scores were computed across all PHI categories.
\\\textbf{Results}: Our model achieved overall F1 scores of 0.973 on the Penn dataset and 0.996 on the Stanford dataset, outperforming or maintaining the previous state-of-the-art model performance. Synthetic PHI evaluation showed consistent detectability (overall F1: 0.959 [0.958–0.960]) across 50 independently de-identified Penn datasets. Our model outperformed all vendor systems on synthetic Penn reports (overall F1: 0.960 vs. 0.632-0.754).
\\\textbf{Discussion}: Large-scale, multimodal training improved cross-institutional generalization and robustness. Synthetic PHI generation preserved data utility while ensuring privacy.
\\\textbf{Conclusion}: A transformer-based de-identification model trained on diverse radiology datasets outperforms prior academic and commercial systems in PHI detection and establishes a new benchmark for secure clinical text processing.

\section{Introduction}
\begin{figure}[h]
  \centering
  \includegraphics[width=0.9\textwidth]{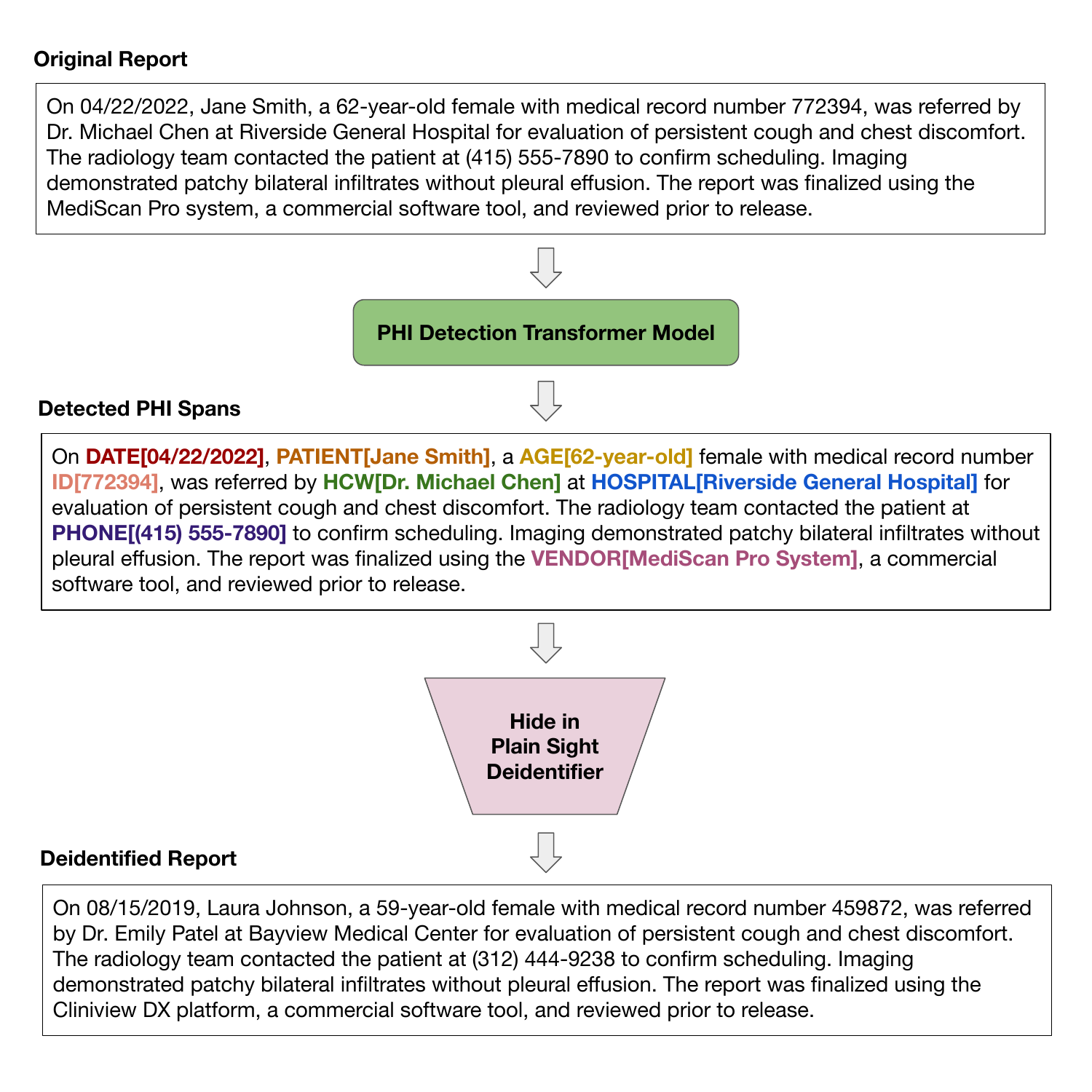}
  \caption{Our pipeline for de-identifying protected health information (PHI) in clinical reports. Each original report is first passed through our transformer model to detect PHI across 8 different classes. Then, the PHI at the detected locations are replaced with realistic synthetic alternatives through the hide-in-plain-sight method. The final result is a de-identified report in which all PHI has been hidden with plausible synthetic substitutes.}
  \label{fig:main}
\end{figure}
The process of de-identifying medical reports entails identifying and excising all protected health information (PHI), as stipulated by the Health Insurance Portability and Accountability Act of 1996 (HIPAA). HIPAA specifies multiple PHI categories, including dates, names, geographic identifiers, and phone numbers. Although the primary aim of HIPAA is to safeguard patient privacy by limiting access to PHI-containing data, such data are vital for training machine learning models designed to tackle text-based medical challenges. Automating the de-identification of medical documents is thus crucial for promoting the development of ML techniques and instruments.

Previous work \cite{pierrechambon} developed a state-of-the-art automated de-identification pipeline for radiology reports that detects PHI entities and replaces them with realistic surrogates, effectively "hiding in plain sight." This PHI detection model achieved an F1 score of 97.9 on radiology reports from a known institution and 99.6 from a new institution, indicating that a transformer-based de-identification pipeline can achieve state-of-the-art performance on radiology reports and other medical documents.

As outlined in Figure \ref{fig:main}, the goal of this project is to improve upon the state-of-the-art transformer-based de-identification pipeline by fine-tuning the model on the combined CheXpert Plus \cite{chexpertplus} and RadGraph-XL \cite{rgxl} datasets from Stanford University, which introduces the AGE PHI category to the base transformer model and integrates two new radiology examination types (brain MRI and abdomen/pelvis CT) in addition to chest radiographs and chest CTs used during training. We benchmark our model for free-text radiology report PHI detection on test sets from both the University of Pennsylvania \cite{penn} and Stanford \cite{chexpertplus, rgxl} and find that we outperform or maintain the state-of-the-art model performance across 8 PHI categories. We also compare the PHI detection performance of our modal to publicly available cloud vendor solutions, leveraging the APIs of Google Cloud Platform (GCP) \cite{gcp_sensitive_data}, Azure Health De-identification Service (Azure) \cite{azure_deid}, and Amazon Comprehend Medical (AWS) \cite{aws_comprehend_medical}, and find that our model similarly outperforms these techniques across its detectable PHI categories.

\subsection{Related Work}

As outlined in previous work \cite{pierrechambon}, de-identification methodologies in automated systems are generally categorized into four types: rule-based, machine learning (ML), hybrid, and deep learning (DL) approaches. 

Rule-based systems \cite{Neamatullah2008, Kayaalp2014} function through the pattern-matching of textual elements, boasting simplicity in their implementation, clarity in interpretation, and ease of modification. However, they are difficult to construct, often fail to generalize to new, unseen data, and struggle with linguistic anomalies such as misspellings and abbreviations.

Machine learning systems treat de-identification as a sequence labeling challenge, producing label predictions for a sequence of input tokens. These systems are adept at discerning complex patterns that may elude human detection. While rule-based algorithms have traditionally dominated PHI detection automation, the advent of ML has seen a progression from conditional random fields to long short-term memory (LSTM) networks \cite{dernoncourt-etal-2017-neuroner, Dernoncourt2017}, and most recently, to transformer architectures \cite{Johnson2020}. Transformers, in particular, have set the new standard for state-of-the-art performance across numerous natural language processing (NLP) tasks \cite{devlin2019bert,vaswani2023attention}. The actual process of PHI removal or alteration has not been as extensively researched. However, the 'hide in plain sight' \cite{Carrell2013} strategy offers a more robust and promising avenue for maintaining the utility of de-identified text while protecting patient privacy. 

Recently, Kocaman et al. \cite{kocaman2023accuracy} introduced a hybrid context-based model architecture for PHI detection in medical records, leveraging transformer architectures. Their findings indicate that this model significantly outperforms traditional machine learning methods and rule-based systems, particularly in managing diverse and complex datasets. Moreover, several cloud computing platforms \cite{gcp_sensitive_data, aws_comprehend_medical} have released automated PHI removal tools, further advancing the field of de-identification. Limited information is available on the relative performance of these methods.

Our study involves the evaluation of the efficacy of various de-identification systems in processing free-text radiology reports. Google Cloud's solution can detect up to 150 PHI entities and allows users to define custom entity detectors, providing a flexible approach to de-identification through methods like redaction, replacement with specified surrogate values, masking, cryptography-based solutions, and tokenization. Azure Health De-identification Service is capable of identifying 27 PHI entities with surrogation. Amazon Comprehend Medical detects 8 PHI entities but does not handle data de-identification within its service. Our proprietary system is designed to detect 8 PHI entities, adding the AGE PHI category to the state-of-the-art transformer model, in addition to a larger-scale training set that includes a new institution and two new radiology examination types (abdomen/pelvis CT and brain MRI). Following the state-of-the-art approach, we utilize a "hiding in plain sight" methodology for synthetic PHI generation. Specifically, our approach deploys a rule-based generator that replaces each detected PHI span with a synthetic one. This comparison of features and capabilities across different platforms highlights the strengths and limitations inherent in each solution, guiding the selection process for institutions prioritizing certain aspects of de-identification over others.

\section{Methods}

\subsection{Data Collection and Annotation}
\begin{table}[h!]
\centering
\begin{tabular}{l r r r}
\hline
\textbf{Category} & \textbf{Stanford Train} & \textbf{Stanford Test} & \textbf{Penn Test} \\
\hline
Total PHI tokens     & 3,412,779  & 854,189   & 6,386 \\
Total non-PHI tokens & 29,184,969 & 7,283,841 & 260,110 \\
\hline
\multicolumn{4}{l}{\textbf{PHI Class}} \\
\hline
AGE      & 4,186   & 1,033   & -- \\
DATE     & 2,055,204 & 513,473 & 5,028 \\
HCW      & 267,881 & 67,693  & 760  \\
HOSPITAL & 1,360   & 375     & 229  \\
ID       & 991,784 & 248,096 & 164  \\
PATIENT  & 530     & 106     & 63   \\
PHONE    & 65,411  & 16,463  & 25   \\
VENDOR   & 26,423  & 6,950   & 117  \\
\hline
\end{tabular}
\caption{Token counts for PHI vs. non-PHI categories across Stanford (Train/Test) \cite{chexpertplus, rgxl} and  Penn (Test) \cite{penn} datasets. The PHI categories represented across both datasets include dates (DATE), health care worker names (HCW), hospital names (HOSPITAL), vendor and software names (VENDOR), unique identifiers (ID), patient names (PATIENT), phone numbers (PHONE), and ages (AGE). AGE was not an annotated PHI class in the Penn test set.}
\label{tab:tokens}
\end{table}

The Penn test set \cite{penn} comprises 1,023 radiology reports manually labeled by radiologists from the University of Pennsylvania, with 6,386 total PHI tokens, as shown in Table \ref{tab:tokens}. It extensively contains date-related PHI (DATE), but also includes health care worker names (HCW), hospital names (HOSPITAL), vendor and software names (VENDOR), unique identifiers (ID), patient names (PATIENT), and phone numbers (PHONE), with the PHI token counts per PHI class detailed in Table \ref{tab:tokens}.

We also consolidate two Stanford University radiology datasets enriched with PHI annotations for DATE, HOSPITAL, VENDOR, HCW, PATIENT, PHONE, ID, and age-related PHI (AGE), with the token counts per PHI class detailed in Table \ref{tab:tokens}. CheXpert Plus is a multimodal dataset consisting of 223,228 unique chest X-ray images paired with 187,711 unique free-text reports. RadGraph-XL is a large-scale, text-only corpus of 2,300 radiology reports spanning chest CT, abdomen/pelvis CT, brain MRI, and chest X-ray examination types, with 2,000 of the reports coming from Stanford University. For both datasets, PHI categories were first detected automatically using the state-of-the-art model \cite{pierrechambon} and subsequently corrected through manual clinical expert review to ensure accuracy and completeness. We randomly divide the datasets into an overall 80/20 training/testing split. Due to its temporary unavailability, we were unable to benchmark against the i2b2 de-identification datasets \cite{i2b21, i2b22}.

\subsection{Model Training}
We fine-tuned the publicly available state-of-the-art transformer model \cite{pierrechambon} for token-level PHI detection on our consolidated Stanford training dataset. The model follows the architecture of the biomedical BERT model PubMedBERT \cite{pubmedbert}, with a linear token-level classification head. We extended the model to include an additional AGE label by modifying the classification head while preserving pre-trained weights for existing categories. During training, due to input token limits, we chunked reports into a maximum of 512 tokens each, ensuring splits only took place between sentences. We used a weighted cross-entropy loss to up-weight PHI classes by a factor of 3. Our model was trained for 5 epochs, using a learning rate of 1e–6, gradient accumulation over 8 steps, and a batch size of 1 per device.
\subsection{Experiments}
We designed three experiments to comprehensively evaluate our de-identification model. 

\textbf{Experiment 1: Benchmarking Against Original Model}  
We first benchmarked our model against the previously published state-of-the-art transformer-based de-identification model \cite{pierrechambon}. This evaluation was performed at the token level on the Penn test set \cite{penn} and our Stanford test set \cite{chexpertplus, rgxl}. We measured precision, recall, and F1 across all PHI categories shared between the models, with AGE evaluated only for our system. The ground-truth annotations provided in the datasets served as the basis for evaluation.  

\textbf{Experiment 2: Synthetic PHI Generation Quality}  
We evaluated the stability and fidelity of our synthetic PHI generation. We applied our model to the Penn test set to first detect PHI entities, and then used the hide-in-plain-sight method to replace each detected PHI with a realistic synthetic surrogate. This process was repeated 50 times to create 50 independently de-identified versions of the same corpus. Each synthetic dataset was then reprocessed with our model, and performance metrics were computed against the synthetic PHI labels produced during generation. We report mean precision, recall, and F1 scores with 95\% confidence intervals across these datasets.  
\begin{table}[h!]
\centering
\begin{tabular}{l l l l}
\hline
\textbf{Our Category} & \textbf{GCP Mapping} & \textbf{AWS Mapping} & \textbf{Azure Mapping} \\
\hline
DATE     & DATE                & DATE             & Date \\
ID       & GENERIC\_ID         & ID               & MedicalRecord / IDNum \\
PHONE    & PHONE\_NUMBER       & PHONE\_OR\_FAX    & Phone \\
PATIENT  & PERSON\_NAME        & NAME             & Patient \\
HCW      & PERSON\_NAME        & NAME             & Doctor \\
HOSPITAL & LOCATION            & ADDRESS          & Hospital \\
VENDOR   & ORGANIZATION\_NAME  & --               & Organization \\
\hline
\end{tabular}
\caption{Entity mapping between our PHI categories and corresponding label sets across GCP, AWS, and Azure de-identification frameworks. A dash (--) indicates that no direct mapping was available.}
\label{tab:mappings}
\end{table}

\textbf{Experiment 3: Benchmarking Against Cloud Vendor Solutions}
We compared our model with commercial cloud vendor systems, including Google Cloud Platform (GCP) \cite{gcp_sensitive_data}, Amazon Comprehend Medical (AWS) \cite{aws_comprehend_medical}, and the Azure Health De-identification Service \cite{azure_deid}. To protect patient privacy, these evaluations were conducted on the first set (of the 50 from Experiment 2) of synthetic reports generated from the Penn dataset. Our model first detected PHI and then applied the hide-in-plain-sight method to generate synthetic surrogates, ensuring no sensitive information was exposed to vendor systems. Vendor predictions were then compared against the synthetic PHI labels generated by our model.  

In order to compare the performance of commercial PHI detection systems with our own model, we normalized their predefined PHI categories to the entity classes used in our evaluation framework. The mappings are as shown in Table \ref{tab:mappings}, and this harmonization of label spaces across Azure, AWS, and GCP ensured a consistent and fair comparison with our model.  

\subsection{Evaluation}
For all experiments, we evaluated models using precision, recall, and F1 score. For Experiment 1, reports exceeding the maximum sequence length were segmented into chunks of at most 512 tokens, with boundaries aligned to sentence breaks. Predictions were reconstructed across segments. Ground-truth labels from the Penn and Stanford datasets were compared directly to predicted outputs.  

For Experiments 2 and 3, the evaluation was performed against the synthetic labels generated by our system during de-identification, reflecting the PHI surrogates inserted via the hide-in-plain-sight method. For the synthetic PHI experiment, we report 95\% confidence intervals across the 50 generated datasets to quantify the variability introduced by surrogate generation. For cloud vendor comparisons, performance was evaluated on the harmonized label space using the synthetic Penn reports, with “-” reported for categories lacking a direct vendor mapping.

\section{Results}

\subsection{Comparative Performance Analysis of Our Model in PHI Detection on Real Data}
\begin{table}[h!]
\centering
\begin{subtable}{\textwidth}
\centering
\begin{tabular}{lcccccc}
\hline
\multirow{2}{*}{\textbf{PHI Class}} & \multicolumn{3}{c}{\textbf{Original Model}} & \multicolumn{3}{c}{\textbf{Our Model}} \\
\cline{2-7}
 & Precision & Recall & F1 & Precision & Recall & F1 \\
\hline
Overall   & 0.979 & 0.975 & 0.977 & 0.972 & 0.974 & 0.973 \\
DATE      & 0.991 & 0.986 & 0.989 & 0.992 & 0.983 & 0.987 \\
ID        & 0.964 & 0.982 & 0.973 & 0.943 & 1.00 & 0.970 \\
HCW       & 0.943 & 0.984 & 0.963 & 0.931 & 0.982 & 0.956 \\
HOSPITAL  & 0.913 & 0.873 & 0.893 & 0.938 & 0.852 & 0.892 \\
PATIENT   & 1.00 & 0.921 & 0.959 & 1.00 & 0.921 & 0.959 \\
PHONE     & 0.808 & 0.840 & 0.824 & 0.781 & 1.00 & 0.877 \\
VENDOR    & 0.811 & 0.658 & 0.726 & 0.664 & 0.795 & 0.724 \\
\hline
\end{tabular}
\caption{Performance on the Penn test set \cite{penn} for PHI detection. AGE is not annotated in this dataset.}
\end{subtable}

\vspace{1em}

\begin{subtable}{\textwidth}
\centering
\begin{tabular}{lcccccc}
\hline
\multirow{2}{*}{\textbf{PHI Class}} & \multicolumn{3}{c}{\textbf{Original Model}} & \multicolumn{3}{c}{\textbf{Our Model}} \\
\cline{2-7}
 & Precision & Recall & F1 & Precision & Recall & F1 \\
\hline
Overall   & 0.996 & 0.989 & 0.993 & 0.999 & 1.00 & 0.996 \\
DATE      & 0.999 & 0.999 & 0.999 & 1.00 & 1.00 & 1.00 \\
ID        & 0.998 & 0.997 & 0.997 & 1.00 & 1.00 & 1.00 \\
HCW       & 0.996 & 0.992 & 0.994 & 0.999 & 1.00 & 0.999 \\
HOSPITAL  & 0.164 & 0.693 & 0.265 & 0.920 & 0.853 & 0.885 \\
PATIENT   & 0.188 & 0.821 & 0.306 & 0.875 & 0.793 & 0.832 \\
PHONE     & 0.993 & 0.996 & 0.995 & 1.00 & 0.997 & 0.999 \\
VENDOR    & 0.961 & 0.166 & 0.283 & 0.989 & 0.995 & 0.992 \\
AGE       &   --  &   --  &   --  & 0.985 & 0.976 & 0.981 \\
\hline
\end{tabular}
\caption{Performance on our Stanford test set \cite{chexpertplus, rgxl} for PHI detection. Original model \cite{pierrechambon} does not predict AGE.}
\end{subtable}

\caption{Comparison of precision, recall, and F1 scores between the original state-of-the-art model \cite{pierrechambon} and our fine-tuned model for PHI detection across two test sets.}
\label{tab:results}
\end{table}
Table~\ref{tab:results} presents results comparing our fine-tuned model with the previous state-of-the-art model \cite{pierrechambon} on the Penn and our Stanford test sets. On the Penn dataset, our model largely maintained the high performance of the original model across all PHI categories. For example, F1 scores for DATE (0.987 vs. 0.989) and ID (0.970 vs. 0.973) were nearly identical between models, while PATIENT performance was unchanged (0.959). Our model achieved modest improvement over the original system in PHONE (F1: 0.877 vs. 0.824), reflecting an enhanced robustness in this category. These findings indicate that our fine-tuned system preserves the strengths of the baseline model while providing modest incremental gains.

On our Stanford test set, our model substantially outperformed the original system. The overall F1 score improved from 0.993 to 0.996, with gains observed across nearly all categories. Notably, our model achieved perfect F1 scores for DATE, ID, and PHONE, and showed substantial improvements in difficult categories such as HOSPITAL (F1: 0.885 vs. 0.265) and PATIENT (0.832 vs. 0.306). Additionally, the introduction of the AGE category yielded high performance (F1: 0.981), highlighting the benefit of extending the model to new PHI types. These results confirm that large-scale training with diverse Stanford data enhances generalization and yields improved performance in real-world evaluation settings.

\subsection{Evaluating Synthetic PHI Generation by Our Model}
\begin{table}[h!]
\centering
\begin{tabular}{lccc}
\hline
\textbf{PHI Class} & Precision & Recall & F1 \\
\hline
Overall   & 0.960 [0.959–0.961] & 0.959 [0.958–0.960] & 0.959 [0.958–0.960] \\
DATE     & 0.985 [0.984–0.986] & 0.987 [0.986–0.988] & 0.986 [0.985–0.987] \\
HCW       & 0.947 [0.942–0.953] & 0.948 [0.943–0.954] & 0.947 [0.942–0.953] \\
HOSPITAL  & 0.805 [0.790–0.820] & 0.782 [0.766–0.797] & 0.789 [0.774–0.804] \\
PATIENT   & 0.965 [0.949–0.981] & 0.980 [0.967–0.993] & 0.969 [0.954–0.983] \\
UNIQUE    & 0.852 [0.832–0.872] & 0.882 [0.863–0.902] & 0.860 [0.840–0.879] \\
PHONE     & 0.972 [0.954–0.991] & 0.974 [0.956–0.993] & 0.973 [0.955–0.992] \\
VENDOR    & 0.806 [0.786–0.825] & 0.865 [0.847–0.883] & 0.826 [0.807–0.844] \\
\hline
\end{tabular}
\caption{PHI detection performance with 95\% confidence intervals of our model on 50 de-identified versions of the Penn dataset}
\label{tab:synth}
\end{table}
We evaluated the stability and fidelity of our model’s synthetic PHI generation using the Penn dataset, as seen in Table \ref{tab:synth}. Across 50 independently de-identified versions of the corpus, our model consistently achieved high scores, with narrow confidence intervals. The overall F1 was 0.959 [0.958–0.960], and strong performance was observed across all categories, including DATE (0.986 [0.985–0.987]), PATIENT (0.969 [0.954–0.983]), and PHONE (0.973 [0.955–0.992]). Performance remained robust even in more variable categories such as VENDOR (0.826 [0.807–0.844]) and HOSPITAL (0.789 [0.774–0.804]). These results demonstrate that the hide-in-plain-sight method reliably produces realistic surrogate PHI while preserving detectability across multiple de-identification passes. The stability of these results supports the utility of synthetic corpora for downstream evaluation and benchmarking without exposing real PHI.

\subsection{Comparative Performance Analysis of Our Model Versus Cloud Vendors on Synthetic Data}
\begin{table}[h!]
\centering
\begin{tabular}{lcccc}
\hline
\textbf{Class} & \textbf{Model} & Precision & Recall & F1 \\
\hline
Overall   & Our Model & 0.962 & 0.958 & 0.960 \\
          & GCP       & 0.553 & 0.737 & 0.632 \\
          & AWS       & 0.736 & 0.773 & 0.754 \\
          & Azure     & 0.704 & 0.795 & 0.747 \\
\hline
DATE      & Our Model & 0.989 & 0.991 & 0.990 \\
          & GCP       & 0.815 & 0.884 & 0.848 \\
          & AWS       & 0.864 & 0.911 & 0.887 \\
          & Azure     & 0.868 & 0.946 & 0.905 \\
\hline
ID        & Our Model & 0.902 & 0.931 & 0.912 \\
          & GCP       & 0.546 & 0.162 & 0.250 \\
          & AWS       & 0.462 & 0.487 & 0.474 \\
          & Azure     & 0.758 & 0.694 & 0.725 \\
\hline
HCW       & Our Model & 0.934 & 0.935 & 0.934 \\
          & GCP       & 0.118 & 0.306 & 0.171 \\
          & AWS       & 0.211 & 0.202 & 0.207 \\
          & Azure     & 0.137 & 0.145 & 0.141 \\
\hline
HOSPITAL  & Our Model & 0.813 & 0.802 & 0.806 \\
          & GCP       & 0.113 & 0.141 & 0.125 \\
          & AWS       & 0.173 & 0.219 & 0.193 \\
          & Azure     & 0.374 & 0.531 & 0.439 \\
\hline
PATIENT   & Our Model & 0.929 & 1.00 & 0.952 \\
          & GCP       & 0.118 & 0.306 & 0.171 \\
          & AWS       & 0.211 & 0.202 & 0.207 \\
          & Azure     & 0.467 & 0.636 & 0.539 \\
\hline
PHONE     & Our Model & 1.00 & 1.00 & 1.00 \\
          & GCP       & 1.00 & 0.667 & 0.800 \\
          & AWS       & 0.800 & 0.667 & 0.727 \\
          & Azure     & 0.111 & 0.333 & 0.167 \\
\hline
VENDOR    & Our Model & 0.761 & 0.790 & 0.771 \\
          & GCP       & 0.036 & 0.128 & 0.056 \\
          & AWS       &   --  &   --  &   --  \\
          & Azure     & 0.071 & 0.026 & 0.038 \\
\hline
\end{tabular}
\caption{Comparison of PHI detection performance across our model, GCP, AWS, and Azure.}
\label{tab:cloud}
\end{table}
We benchmarked our system against GCP, AWS, and Azure using synthetic versions of the Penn dataset (Table~3). Our model achieved the highest overall performance with an F1 of 0.960, surpassing GCP (0.632), AWS (0.754), and Azure (0.747). Performance differences were particularly pronounced in several categories. For example, our model achieved near-perfect detection of DATE (F1: 0.990) compared to 0.848 for GCP, 0.887 for AWS, and 0.905 for Azure. For ID, our model reached 0.912 versus substantially lower scores for GCP (0.250), AWS (0.474), and Azure (0.725). The gap was even larger for PATIENT and HCW categories, where our model achieved F1 scores of 0.952 and 0.934, while the best competing vendor reached only 0.539 and 0.207, respectively. Notably, our model was the only system to achieve near perfect performance on PHONE (F1: 1.00), compared to 0.800 for GCP, 0.727 for AWS, and 0.167 for Azure. For VENDOR, cloud services struggled significantly, with GCP and Azure achieving 0.056 and 0.038 F1 score respectively, while AWS lacked a corresponding label.

\section{Discussion and Conclusion}
\begin{figure}[h]
  \centering
  \begin{subfigure}[b]{0.45\textwidth}
    \centering
    \includegraphics[width=\textwidth]{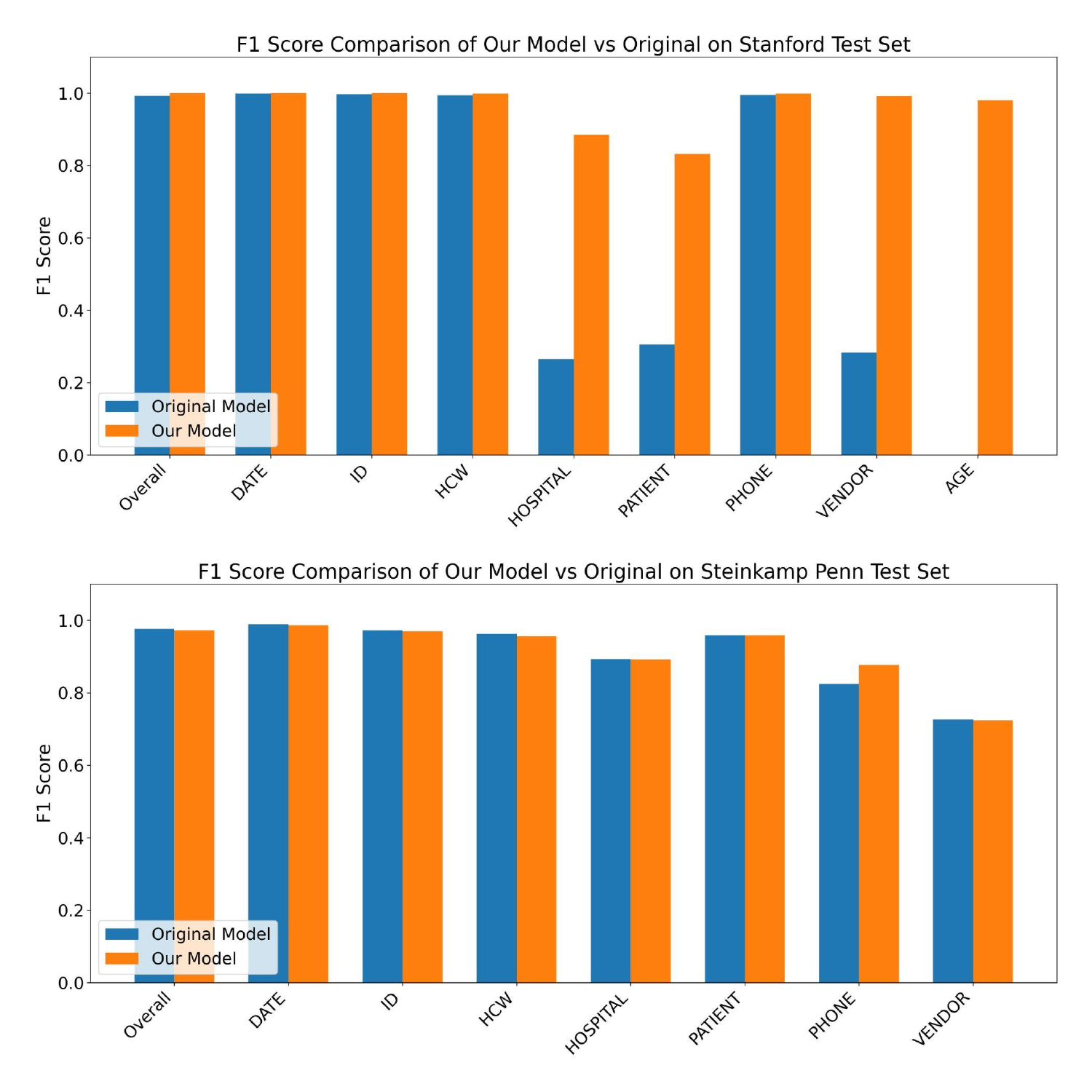}
    \caption{Comparative performance of our model on real data.}
  \end{subfigure}
  \hfill
  \begin{subfigure}[b]{0.45\textwidth}
    \centering
    \includegraphics[width=\textwidth]{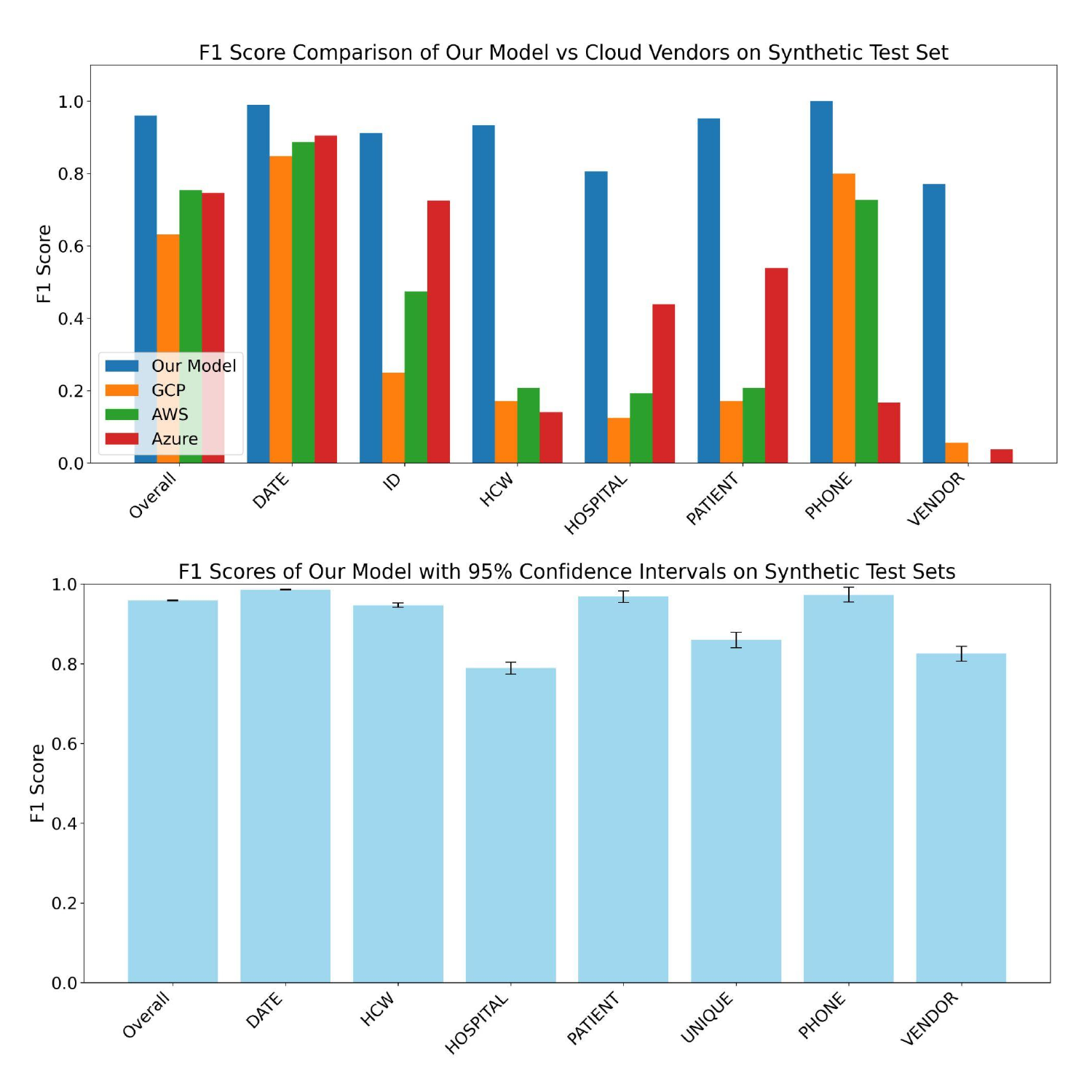}
    \caption{Comparative performance of our model on synthetic data.}
  \end{subfigure}
  
  \caption{Overall summary of our model performance across real and synthetic data benchmarked against competitors.}
  \label{fig:summary}
\end{figure}

Across three experiments, summarized in Figure \ref{fig:summary}, our results demonstrate that large-scale training on diverse Stanford datasets enables our model to maintain state-of-the-art PHI detection performance on the Penn test set while achieving substantial improvements on the Stanford test set. On Penn, our model matched the original state-of-the-art system across most PHI categories while delivering incremental improvements for the PHONE entity. On the Stanford test set, our system consistently outperformed the baseline model across nearly all categories, most notably for HOSPITAL and PATIENT. The addition of the AGE category also demonstrated strong performance, highlighting the benefits of incorporating new PHI types. The Stanford dataset also introduced two new radiology examination types (i.e. abdomen/pelvis CT and brain MRI) into training. This broader examination coverage likely contributed to the improved generalizability of our model compared to the baseline.

The second experiment confirmed that our approach to synthetic PHI generation via the hide-in-plain-sight method produces stable and realistic surrogate data. Across 50 independently de-identified versions of the Penn dataset, performance remained consistent, with narrow confidence intervals across all PHI categories. These results validate the fidelity of our surrogate generation process, showing that synthetic corpora can serve as reliable stand-ins for ground-truth labels in privacy-preserving evaluations. This is particularly important for enabling safe benchmarking against third-party systems without exposing sensitive patient data.

In the third experiment, our model demonstrated superior performance over publicly available cloud vendor systems, including GCP, AWS, and Azure, when evaluated on synthetic Penn reports. Our model consistently outperformed all three vendors across nearly every PHI category, achieving the highest overall F1 score of 0.960 compared to 0.632 for GCP, 0.754 for AWS, and 0.747 for Azure. Performance gaps were most pronounced in PATIENT, HCW, and VENDOR detection, categories where vendors struggled significantly but our model maintained strong results. These results highlight the limitations of current vendor solutions for comprehensive PHI detection in radiology reports, and underscore the advantage of domain-specific models trained on large, diverse medical datasets.

While these findings are promising, several limitations should be acknowledged. First, for cloud vendor comparisons, performance was measured against synthetic labels generated by our system rather than ground-truth annotations. Although this approach was necessary to protect patient privacy, it means that vendor results were evaluated relative to our model’s labeling decisions. Second, our evaluation focused on radiology reports from two institutions, and further work is needed to assess generalizability to other clinical document types and healthcare settings. Finally, although our model introduces new PHI categories and modalities, additional categories specified under HIPAA (e.g. geographic identifiers beyond hospitals) remain unaddressed.

In summary, our study demonstrates that scaling transformer-based de-identification models with large, multimodal clinical datasets enables both the preservation of existing state-of-the-art performance and significant improvements on challenging categories and test sets from multiple institutions. Our model not only generates reliable synthetic PHI for safe evaluation but also outperforms publicly available cloud vendor solutions, establishing a new benchmark for free-text radiology report de-identification. Expanding to broader clinical note types, integrating additional PHI categories, and exploring data from more institutions are promising next steps in further strengthening automated de-identification systems.

\section{Acknowledgements}
This work was supported in part by the National Institute of Biomedical Imaging and Bioengineering (NIBIB) of the National Institutes of Health under contract 75N92020C00021 and through The Advanced Research Projects Agency for Health (ARPA-H).

\newpage
\bibliographystyle{plain}
\bibliography{sample}

\end{document}